\documentclass[letterpaper]{article} 
\usepackage{aaai2026}  
\usepackage{times}  
\usepackage{helvet}  
\usepackage{courier}  
\usepackage[hyphens]{url}  
\usepackage{graphicx} 
\usepackage{natbib}  
\usepackage{caption} 
\usepackage{algorithm}
\usepackage{algorithmicx}
\usepackage{algpseudocode}
\usepackage{amsmath,amsfonts}
\usepackage{newfloat}
\usepackage{listings}
\usepackage{booktabs}
\usepackage[table,dvipsnames]{xcolor}
\usepackage{pifont}
\newcommand{\cmark}{\ding{51}}
\newcommand{\xmark}{\ding{55}}
\DeclareCaptionStyle{ruled}{labelfont=normalfont,labelsep=colon,strut=off} 
\floatstyle{ruled}
\newfloat{listing}{tb}{lst}{}
\floatname{listing}{Listing}
\title{Evolutionary System 2 Reasoning: An Empirical Proof}
\author{
    Zeyuan Ma\textsuperscript{\rm 1}\textsuperscript{\rm 4},
    Wenqi Huang\textsuperscript{\rm 1},
    Guo-Huan Song\textsuperscript{\rm 2}\textsuperscript{\rm 3},
    Hongshu Guo\textsuperscript{\rm 1}\textsuperscript{\rm 4},\\
    Sijie Ma\textsuperscript{\rm 1},
    Zhiguang Cao\textsuperscript{\rm 5},
    Yue-Jiao Gong\textsuperscript{\rm 1}\thanks{Corresponding author~(gongyuejiao@gmail.com)}
}
\affiliations{
    \textsuperscript{\rm 1}South China University of Technology, 
    \textsuperscript{\rm 2}Zhejiang Normal University, 
    \textsuperscript{\rm 3}Northern Computility,\\
    \textsuperscript{\rm 4}Panorama Optimization, 
    \textsuperscript{\rm 5}Singapore Management University

}

\begin{document}
\twocolumn[{%
\renewcommand\twocolumn[1][]{#1}%
\maketitle
\begin{center}
    \centering
    \includegraphics[width=0.9\linewidth]{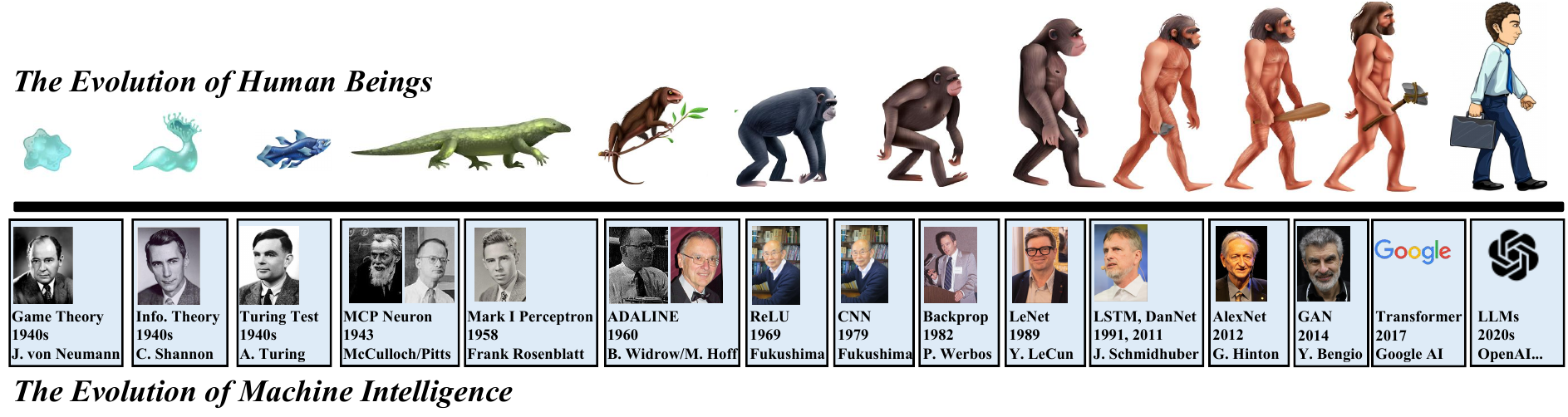}
    \captionof{figure}{An intuitive comparison between the evolution paths of human beings and machine intelligence.}
    \label{fig:intro}
\end{center}
}]

\begin{abstract}
Machine intelligence marks the ultimate dream of making machines' intelligence comparable to human beings. While recent progress in Large Language Models (LLMs) show substantial \emph{specific skills} for a wide array of downstream tasks, they more or less fall shorts in \emph{general intelligence}. Following correlation between intelligence and system 2 reasoning (slow thinking), in this paper, we aim to answering a worthwhile research question: could machine intelligence such as LLMs be evolved to acquire reasoning ability (not specific skill) just like our human beings? To this end, we propose evolutionary reasoning optimization~(ERO) framework which performs \emph{survival of the fittest} over a population of LLMs to search for individual with strong reasoning ability. Given a reasoning task, ERO first initializes multiple LLMs as a population, after which an evolutionary strategy evolves the population to maximize quantified reasoning score of the best individual. Based on experiments on representative testsuites, we claim two surprising empirical discoveries: i) the latest LLMs such as GPT-5 still show limited system 2 reasoning ability; ii) with simple evolution-loop of ERO, a relatively weak model (Qwen-7B) could be enhanced to emerge powerful reasoning ability. Our project can be accessed at \textcolor{blue}{\url{https://github.com/MetaEvo/ERO}} for reproduction needs.     
\end{abstract}


\begin{quote}
    \emph{This paper does not advertise for LLMs, but explores more possibilities.}  \quad \quad \quad \quad \quad \quad \quad  ---\hspace{0.05pt} \emph{The authors}
\end{quote}

\renewcommand{\thefootnote}{\fnsymbol{footnote}}
\footnotetext{*Corresponding author~(gongyuejiao@gmail.com)}

\renewcommand{\thefootnote}{\arabic{footnote}}

\section{Introduction}
Machine intelligence~(often interchangeably used with AI) has experienced ups and downs within a long river of history~\cite{mi-book,ai-book,dl-book}. Since the initial proposal of AI at 1950s~\cite{ai-proposal}, an evolution path has been observed: from basic theories~\cite{shannon,turing} to concrete architectures~\cite{perceptron,cnn,lstm,transformer,mamba} and algorithms~\cite{sgd,backprop,rmsprop,adamw}. Today, the application of AI has spread to every corner of the world. Domains such as image processing~\cite{cv}, nature language processing~\cite{nlp} and scientific discovery~\cite{alphafold} benefit from its learning power and corresponding human-competitive performance.

However, we should not overlook the dark side of advanced machine intelligence (i.e., LLMs) simply due to its twinkling academic and engineering achievements~\cite{llm-app1,llm-app2,llm-app3}. In other words, we have to realize that LLMs, though pre-trained with massive human knowledge prior, may still operate at the pattern recognition~(fast thinking, System 1 reasoning) level, and hence lacks long-chain, deep, logical reasoning ability~(slow thinking, System 2 reasoning), as testified in recent competitions\footnote{\textcolor{blue}{https://arcprize.org/leaderboard}}. 

As illustrated in Figure~\ref{fig:intro}, such System 2 reasoning inability potentially roots from the essential difference between the evolution of machine intelligence and that of our human beings~\cite{evolution-1,evolution-2}. For human beings, we are continually involved in evolutionary process under open-ended environmental selection pressure, which follows the \emph{survival of the fittest} principle proposed by Darwin~\cite{darwin}. The ``open-ended'' term is used to reference extreme generalization scenario where environmental uncertainty is naturally unknown by human beings~\cite{nfl2}. In contrast, almost all machine intelligence instances are trained for specific application scopes explicitly restricted by their developers (human beings). The feedback or learning signal in their learning loops may inherently restricts them from general intelligence with logic reasoning~\cite{nfl1}. To make this point clearer, we borrow the valuable perspective from  developmental psychology~\cite{psy-1}, which holds the position that: human-level intelligence shows generalization and open-endedness and is capable of expanding far beyond its evolution path. More importantly, human is born with innate and evolution-driven knowledge priors such as elementary physics, goal-directness, arithmetic and geometry. These priors enable us to acquire certain skills efficiently~\cite{arc-1}, by System 2 slow thinking.

The gap between existing LLMs and general System 2 reasoning ability motivates us to explore possible solutions. An intuitive yet under-explored thought would be: Given that existing advanced LLMs have absorbed massive knowledge priors through pre-training with internet-scale corpus, can we further evolve them~(e.g., Neuroevolution~\cite{ne}) to attain System 2 reasoning ability? To answer this research question empirically, we in this paper propose \textbf{Evolutionary Reasoning Optimization~(ERO)} framework that enables human-like evolution process for LLMs to adapt themselves in complex tasks that require System 2 reasoning. In our framework, the neural network parameters of a LLM are regarded as a holistic genotype space. At the beginning, given a complex reasoning task, a population of LLMs are randomly born via sampling from the genotype space. Then a ($\mu$+$\lambda$) Evolutionary Strategy~(ES)~\cite{es-1,es-2} is applied to guide the LLM population toward more powerful System 2 reasoning performance on the target task. The evolution rule in our framework is purely objective-oriented: the LLM individual with higher reasoning ability survives and contributes to the reproduction of offspring, which is closely analogous to evolution of human beings. We provide an intuitive illustration in Figure~\ref{fig:evolution} to showcase how ERO evolves a weak Qwen-7B model to surpass powerful GPT-5 model on reasoning tasks. We next provide a brief review of related works in Sec.~\ref{sec:related}, elaborate the technical details of \textbf{ERO} in Sec.~\ref{sec:method} and discuss empirical results in Sec.~\ref{sec:exp} respectively.  


\section{Related Works}\label{sec:related}
\subsection{Reasoning in LLMs}
Reasoning ability is regarded as a key for achieving human-level machine intelligence~\cite{system2-survey}. In particular, it relies on logical reasoning and systematic step-by-step thinking to ensure solving effectiveness on complex tasks, which is typically termed as System 2 reasoning. Compared to System 1 reasoning, which features fast, pattern recognition-based decision mapping, System 2 reasoning presents deliberate slow thinking, resulting in concise and rational problem solving via mitigating cognitive biases in System 1 reasoning. While the swift development of LLMs~(e.g., DeepSeek-v3~\cite{deepseek-v3}, GPT-5~\cite{gpt-5}) shows promising results on understanding and performing human-competitive tasks, they may still lack matched cognitive abilities with human beings in complex reasoning tasks~\cite{arc-2}. 

To improve the capability of reasoning LLMs, initial exploration includes Chain-of-Thought~(CoT)~\cite{cot-1,cot-2} and Tree-of-Thought~(ToT)~\cite{tot}, which focus on preparing high-quality, step-by-step and fine-grained supervision data through decomposing the complex reasoning process into chain or tree structure. Given the data scaling difficulty and single-pass reasoning pattern in CoT and ToT, subsequent works further apply Monte Carlo Tree Search~(MCTS) to allow LLMs revisit, reflect and refine their reasoning paths dynamically~\cite{mcts-1,mcts-2,mcts-3}, or self-improvement strategies~\cite{self-improve-1} that bootstrap training data from either iterative self-reflection~\cite{self-improve-2} or rule-based reasoning path augmentation~\cite{self-improve-3}. Beside these data curation designs, the training paradigm itself also plays crucial role in attaining robust reasoning LLMs. Common practice in up-to-date literature leans to reinforcement fine-tuning~(RFT) with output reward modeling~(ORM)~\cite{orm} or process reward modeling~(ORM)~\cite{prm}. The former emphasizes scoring for final answer correctness and the latter pays efforts on fine-grained step-by-step reward labeling. Test time training~(TTT)~\cite{ttt} is also adopted as effective post-training strategy to mitigate reasoning hallucination. For further reading, we suggest these surveys~\cite{system2-survey,reasoning-survey-1,reasoning-survey-2}.    

\subsection{Reasoning LLMs Benchmarks}

While recent advance of LLMs demonstrates that, with large-scale pre-training on massive and diverse corpus, these novel machine intelligence models rival or even surpass human's performance at specific testbeds~\cite{testbed-1,testbed-2,evogit}, more evidences argue that they lack compositional System 2 reasoning ability in solving complex tasks as general intelligence~\cite{arc-evaluate}. To this end, a large body of related benchmarks have been curated to provide objective and challenging reasoning tasks for evaluating reasoning LLMs. According to their concrete task types, we could generally document them as: 1) Olympic-level mathematical reasoning benchmarks~\cite{aime,var-math}; 2) Real world programming challenges summarized from Github~\cite{swe}; 3) Scientific discovery process~\cite{sciench-bench} in physics, chemistry, etc.; 4) Agentic automation workflow tests, e.g., constructing web application from zero~\cite{agent-bench}; 5) Human-level cognitive ability tests~\cite{arc-1,arc-2} that analog IQ examination.

\begin{figure}
    \centering
    \includegraphics[width=0.99\columnwidth]{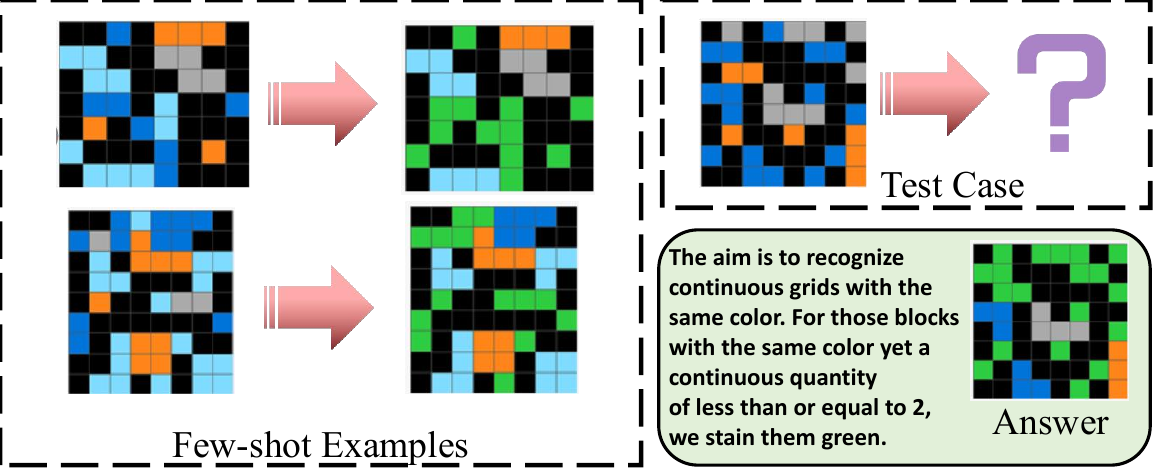}
    \caption{A reasoning task example in ARC benchmark.}
    \label{fig:arc-intro}
\end{figure}

In this paper, we focus on the last benchmark type, of which a representative benchmark is Abstraction and Reasoning Corpus~(ARC) benchmark~\cite{arc-1}. As illustrated in Figure~\ref{fig:arc-intro}, the testing task instance in ARC benchmark includes multiple few-shot examples and a test case for machine intelligence to solve, which stays close in format of psychometric intelligence test~\cite{psy-theory}. To figure out each puzzle, an intelligence must coherently enable its innate prior on object persistence and contact influence, goal-directedness, numbers and counting, etc., just like our human beings. According to the latest leader board~\cite{arc-2}, even the most powerful reasoning-reinforced LLMs~(GPT 5 and Gemini 3) could only achieve scores no more than 50\%, with an evident reasoning gap against human panel~(100\%). This benchmark provides us a desirable testbed.    

\subsection{Evolutionary LLMs Enhancement}
Evolutionary Algorithms~(EAs)~\cite{ec-survey} are meta-heuristics that follow evolutionary principle in nature to optimize given problems through reproduction and selective pressure. Given EAs' high-level alignment with the evolution process of human beings and robust global optimization capability, they have been validated as powerful optimization techniques for many applications~\cite{ec-survey-app}, except LLMs. Recently, initial attempts have been made to explore the possibility of leveraging EAs to enhance LLMs' performances. While limited, these efforts have seen delightful effects such as prompt optimization through textual evolution~\cite{evoprompt}, program evolution through LLM-level genetic programming~\cite{eoh,alphaevo}, novel ability composition through model merge recipes~\cite{mergellm-1,mergellm-2}, incremental and dynamic prompting through evolutionary context engineering~\cite{context-engineer}. However, to the best our knowledge, none of prior works focus on the core vision of LLMs: reasoning like human beings. This highlights the motivation of our paper.

\section{Evolutionary Reasoning Optimization}\label{sec:method}
In this section, we elaborate both the general picture and specific designs of our ERO framework to clarify how we address reasoning enhancement for LLMs via EAs perspective. Generally speaking, ERO operates as a neuroevolution~\cite{ne} approach, which is under the umbrella of evolutionary strategy~(ES)~\cite{es-1,es-2} framework. We present the overall workflow of ERO in Alg.~\ref{alg:algorithm}, where starting from an existing LLM, an iterative searching process is deployed to evolve the parameters of the LLM toward high reasoning performance on the given reasoning task. However, we must note that it is neither practical nor efficient to run ERO in a standard ES procedure. We next detail the key challenges and corresponding tailored designs in ERO.

\noindent\textbf{Sampling Strategy:} In ERO, we have to first determine sampling strategy~(i.e., mean and covariance parameters) to serve as initialization module and hence kick out subsequent evolution process. For the mean parameter $\mu$, we could simply set it as the weights of the LLM~(denoted as $\theta^{0}$). The real challenge is how to determine covariance parameter $\Sigma$. Although ES has been previously used in evolving relatively smaller neural networks~\cite{ne-survey,nes}, where the value ranges of parameters are controllable and hence we could set identical entries for $\Sigma$ matrix, it is absolutely not the case in LLMs. This is backed up by our preliminary experiment, where we conducted a statistical summary on different LLMs and found out that the value ranges of LLMs' parameters vary a lot. However, we also found out that the value ranges of layer-wise parameters are more stable. Based on such observation, we determine the entries of variance matrix $\Sigma$ by the principle below:
\begin{equation}
    \Sigma[k,k] = \epsilon \times \frac{1}{|L_k|}\sum_{n=1}^{|L_k|}\theta^{(0)}[L_k][n]
\end{equation}
where $k$ denotes $k$-th neural network parameters of the selected LLM, $L_k$ is the network layer where $k$-th parameter locate at, $\epsilon$ is value between $0\sim1$ to control the variance strength, $[\cdot]$ is the indexing operation. We set $\Sigma$ once leave it fixed until the end. A population of LLMs with the same architecture with $\theta^{(0)}$ are then sampled by the constructed gaussian distribution~(line 4 in Alg.~\ref{alg:algorithm}).

\begin{figure*}[t!]
    \centering
    \includegraphics[width=0.99\linewidth]{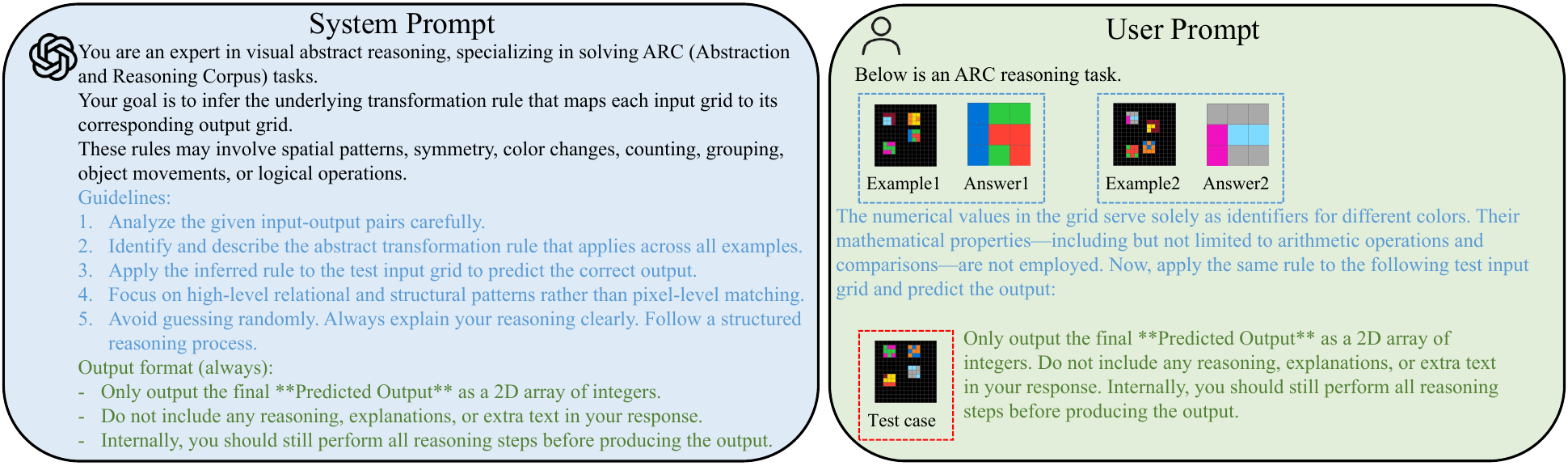}
    \caption{System prompt and User prompt we used across all baselines.}
    \label{fig:prompt}
\end{figure*}

\noindent\textbf{Scoring Function:} Given a population of $\lambda$ sampled LLMs at $g$-th generation: $\{\theta^{(g),i}\}_{i=1}^{\lambda}$, the underlying ES process in ERO needs proper evaluation metric~(scoring function) to measure the reasoning performances of these LLM individuals on the given task $\tau$. A general form of such scoring function can be formulated as $\mathbb{S}(\theta|\tau)$, where $\theta$ denotes a tested LLM. We would like to clarify that our ERO does not restrict concrete implementation of the scoring function, instead, it can be quite flexible to tailor appropriate scoring schemes for different reasoning tasks. One can surely use generic schemes such as process reward model~\cite{prm} that regards any reasoning task as standard reasoning chain and credits those matched reasoning steps. On the other hand, one can also customize special $\mathbb{S}$ function for specific task. Since our ERO is a purely objective-oriented optimization system, all it need is a scalar objective to minimize or maximize. We take the testbed we select for this paper~(ARC benchmark) as an example. In ARC, the answer of a reasoning task instance is typically a 1-D or 2-D array indicating the colors of grids. By representing them as strings, one can simply compute the score as:   
\begin{equation}\label{eq:score}
    \mathbb{S}(\theta|\tau) = 1-\frac{lev(\hat A(\tau|\theta),A(\tau))}{maxLen(A(\tau),\hat A(\tau|\theta))}
\end{equation}
where $lev(\cdot,\cdot)$ is the Levenshtein distance~\cite{edit-distance} between two strings, $A$ and $\hat A$ is the ground truth and predicted answer respectively. In this paper, ERO aims at maximizing the LLM's performance on ARC tasks through maximizing corresponding scoring function values.

\noindent\textbf{Island Architecture:} Given the massive searching space of LLM's neural network parameters, island-based population architecture could be a useful strategy to enhance the searching diversity of underlying ES process, which may further improve the final optimization performance~\cite{island-1}. Besides, since LLMs are inherently aligned with multi-card computational resources and distributed computational methods, island architecture is a coherent choice in LLM-based evolution frameworks~\cite{island-2,island-3}. To this end, our ERO instantiates multiple LLM populations as independent islands, which sample and evaluate LLM individuals~(lines $4\sim5$ of Alg.~\ref{alg:algorithm}) in parallel. The communication~(fitness aggregation) across different islands occurs when we have to aggregate elite LLM individuals and accordingly update the mean and variance parameters of ES process~(lines $6\sim7$ of Alg.~\ref{alg:algorithm}). Unlike vanilla ES, the $\mu$ elite LLM individuals are selected as the $\lfloor \frac{\mu}{Z}\rfloor$ best individuals per island, where $Z$ is the number of islands deployed. Once the elite individuals are voted out, we update the mean parameters used for next-generation sampling by averaged aggregation. Note that we keep a fixed variance matrix $\Sigma$ to maintain continuous exploration strength along the evolution process.

\noindent\textbf{Ray Acceleration:} As we mentioned above, the island architecture allows us incorporate advanced distributed ML techniques to reduce the running complexity of LLM-based evolution frameworks. This is particular useful in our ERO, since the scoring evaluation is actually time-consuming, where each LLM individual is fed with reasoning questions and prompted to output reasoning steps and answers. We hence introduce Ray\footnote{\textcolor{blue}{https://github.com/ray-project/ray}}, a large-scale ML-enabled computational framework, to distribute each island in ERO onto a separate GPU of a multi-GPU computer/cluster. The Ray parallelism not only enables distributed island-based evolution, but also further facilitates fine-grained parallel evaluation within each island, reducing the running time of ERO from days to hours. In practice, the concrete parallel degree varies due to different hardware conditions. 
\begin{table*}[t]
\centering
\resizebox{\textwidth}{!}{
\begin{tabular}{lccccccccccccccc}\toprule
\textbf{Tasks} &
T1 & T2 & T3 & T4 & T5 & T6 & T7 & T8 & T9 & T10 & T11 & T12 & T13 & T14 & T15 \\\midrule

\rowcolor{Gray!20}
\textbf{\emph{Properties}} & & & & & & & & & & & & & & & \\

Object cohesion &
\textcolor{green}{\cmark} & \textcolor{green}{\cmark} & \textcolor{green}{\cmark} & \textcolor{green}{\cmark} & \textcolor{green}{\cmark} & \textcolor{green}{\cmark} & \textcolor{green}{\cmark} & \textcolor{green}{\cmark} &
\textcolor{green}{\cmark} & \textcolor{green}{\cmark} & \textcolor{green}{\cmark} & \textcolor{green}{\cmark} & \textcolor{red}{\xmark} & \textcolor{red}{\xmark} & \textcolor{green}{\cmark} \\

Object persistence &
\textcolor{red}{\xmark} & \textcolor{red}{\xmark} & \textcolor{red}{\xmark} & \textcolor{red}{\xmark} & \textcolor{red}{\xmark} & \textcolor{red}{\xmark} & \textcolor{red}{\xmark} & \textcolor{red}{\xmark} &
\textcolor{red}{\xmark} & \textcolor{red}{\xmark} & \textcolor{red}{\xmark} & \textcolor{red}{\xmark} & \textcolor{red}{\xmark} & \textcolor{green}{\cmark} & \textcolor{red}{\xmark} \\

Object influence via contact &
\textcolor{red}{\xmark} & \textcolor{red}{\xmark} & \textcolor{red}{\xmark} & \textcolor{green}{\cmark} & \textcolor{red}{\xmark} & \textcolor{green}{\cmark} & \textcolor{red}{\xmark} & \textcolor{red}{\xmark} &
\textcolor{green}{\cmark} & \textcolor{red}{\xmark} & \textcolor{red}{\xmark} & \textcolor{green}{\cmark} & \textcolor{red}{\xmark} & \textcolor{red}{\xmark} & \textcolor{red}{\xmark} \\

Goal-directedness &
\textcolor{green}{\cmark} & \textcolor{red}{\xmark} & \textcolor{red}{\xmark} & \textcolor{red}{\xmark} & \textcolor{red}{\xmark} & \textcolor{red}{\xmark} & \textcolor{red}{\xmark} & \textcolor{red}{\xmark} &
\textcolor{red}{\xmark} & \textcolor{red}{\xmark} & \textcolor{red}{\xmark} & \textcolor{red}{\xmark} & \textcolor{red}{\xmark} & \textcolor{red}{\xmark} & \textcolor{red}{\xmark} \\

Numbers and Counting &
\textcolor{red}{\xmark} & \textcolor{red}{\xmark} & \textcolor{red}{\xmark} & \textcolor{red}{\xmark} & \textcolor{red}{\xmark} & \textcolor{green}{\cmark} & \textcolor{green}{\cmark} & \textcolor{green}{\cmark} &
\textcolor{red}{\xmark} & \textcolor{red}{\xmark} & \textcolor{green}{\cmark} & \textcolor{green}{\cmark} & \textcolor{red}{\xmark} & \textcolor{red}{\xmark} & \textcolor{red}{\xmark} \\

Basic Geometry and Topology &
\textcolor{green}{\cmark} & \textcolor{green}{\cmark} & \textcolor{green}{\cmark} & \textcolor{green}{\cmark} & \textcolor{red}{\xmark} & \textcolor{green}{\cmark} & \textcolor{red}{\xmark} & \textcolor{red}{\xmark} &
\textcolor{green}{\cmark} & \textcolor{green}{\cmark} & \textcolor{red}{\xmark} & \textcolor{green}{\cmark} & \textcolor{green}{\cmark} & \textcolor{green}{\cmark} & \textcolor{green}{\cmark} \\

\rowcolor{Gray!20}
\textbf{\emph{Performance}}  & & & & & & & & & & & & & & & \\

\rowcolor{NavyBlue!20}
Ours~(ERO+Qwen2.5-7B) &
\underline{0.7765} & \textbf{0.7820} & \textbf{0.9845} & \underline{0.7828} & \textbf{0.9016} & \textbf{1.0000} & \textbf{0.8100} & \textbf{1.0000} &
\underline{0.9461} & \textbf{0.8182} & \textbf{1.0000} & \underline{0.9627} & \underline{0.7193} & \textbf{0.7315} & \underline{0.7073} \\

\rowcolor{NavyBlue!20}
Qwen2.5-7B &
0.7059 & 0.1132 & 0.9380 & 0.4253 & 0.2623 & 0.9344 & 0.3584 & 0.6400 &
0.8491 & 0.3333 & 0.6400 & 0.9379 & 0.6486 & 0.5370 & 0.6098 \\

Qwen2.5-32B &
0.6118 & 0.3831 & 0.9535 & 0.4143 & 0.5164 & 0.6393 & 0.0924 & \textbf{1.0000} &
0.4315 & 0.4848 & 0.5200 & 0.4752 & \textbf{0.7235} & 0.6205 & 0.4268 \\

GPT-4o-mini &
0.1401 & 0.1200 & 0.9380 & 0.3875 & 0.3115 & 0.9344 & 0.3122 & 0.6400 &
0.9212 & 0.4545 & 0.6400 & 0.9379 & 0.5489 & 0.5507 & 0.6341 \\

GPT-4o &
0.6195 & 0.2699 & \underline{0.9767} & 0.4434 & 0.2295 & 0.7541 & \underline{0.6380} & \textbf{1.0000} &
0.8880 & \underline{0.6061} & 0.6400 & \textbf{0.9689} & 0.6861 & \underline{0.7205} & 0.6341 \\

GPT-5 &
\textbf{0.8647} & \underline{0.6505} & 0.4961 & \textbf{1.0000} & \underline{0.8934} & \textbf{1.0000} & 0.4027 & \textbf{1.0000} &
\textbf{0.9647} & 0.5455 & \underline{0.8200} & 0.9472 & 0.4376 & 0.3260 & \textbf{0.7195} \\ 
\bottomrule
\end{tabular}
}
\caption{Pass@1 scores of LLMs baselines across 15 ARC tasks, with their task properties attached at the top of the table.}
\label{tab:results}
\end{table*}

\begin{algorithm}[t]
\caption{Evolutionary Reasoning Optimization}
\label{alg:algorithm}
\textbf{Input}: LLM $\theta^{(0)}$; reasoning task $\tau$; population size $\lambda$; elite group size $\mu$; optimization budget \emph{G}.\\
\textbf{Output}: best LLM individual $\theta^*$ found ever.
\begin{algorithmic}[1] 
\State Attain layer-wise covariance $\Sigma$ from $\theta^{(0)}$
\State Let $g=1$
\While{$g \leq G$}
\State Sample $\lambda$ LLMs: $\{\theta^{(g),i}\}_{i=1}^{\lambda}\sim \mathcal{N}(\theta^{(g-1)}, \Sigma)$
\State Evaluate their reasoning scores: $\{\mathbb{S}(\theta^{(g),i}|\tau)\}_{i=1}^{\lambda}$
\State Select $\mu$ top-scoring LLMs: $\{\hat \theta^{(g),j}\}_{j=1}^{\mu}$
\State Update $\theta^{(g)} = \frac{1}{\mu}\sum_{j=1}^{\mu} \hat\theta^{(g),j}$
\State $g = g+1$
\EndWhile
\State \textbf{return} the LLM individual with the best score
\end{algorithmic}
\end{algorithm}

\noindent\textbf{Cache Optimization:} One may question about how could a large population of LLMs be loaded within a single 4-GPU or 8-GPU computer/cluster, since a single LLM may require at least $10\sim20$ GB GPU memory. The solution we propose is to subtly and flexibly leverage limited cache memory. In specific, we only maintain necessary LLM information in an \emph{on-the-fly} fashion for each island (i.e., each GPU node). The necessary LLM information includes the mean parameters at current optimization generation~($\theta^{(g)}$), the layer-wise variance matrix $\Sigma$, the elite pool used for maintaining $\lfloor \frac{\mu}{Z}\rfloor$ elite LLM individuals. The elite pool is dynamically updated when a newly sampled LLM individual gets better reasoning score than those in the pool, where the older elite is in-place replaced by the newly sampled one. With such cache memory optimization, ERO could evolve hundreds of LLM individuals simultaneously on GPU memory-limited platform.

\section{Empirical Validation and Discussion}\label{sec:exp}
\subsection{Experimental Setup}
We list detailed settings of each parts in ERO here, which could be generally divided into three categories:

\noindent\textbf{ERO's Settings:} We select \emph{Qwen-7B}\footnote{\textcolor{blue}{https://huggingface.co/Qwen/Qwen2.5-VL-7B-Instruct}} as the initial LLM $\theta^{(0)}$ to be evolved. The reason behind such selection is that this relatively poor-reasoning model could facilitate validation on effectiveness of our ERO. For the hyper-parameters of the underlying island-based ES, we set its population size $\lambda=1000$ which are evenly distributed to $Z=4$ islands, elite pool size $\mu=4$ and the optimization budget $G=12$ generations. All experiments are run on a high-performance instance of a GPU cluster, which comprises an Intel Xeon 8558P CPU, 128 GB RAM and 4$\times$64 GB virtual GPU nodes based on Nvidia H20 GPU. 

\noindent\textbf{Testbed:} As a preliminary study and due to limited computational resources, in this paper, we have randomly sampled 15 reasoning task instances from hundreds of instances in ARC-1 benchmark~\cite{arc-1}. We mark these 15 tested instances as T1$\sim$T15. We present at upper half of Table~\ref{tab:results} the fine-grained properties of these instances in terms of their correspondence to innate cognitive abilities of human beings. Refer to our project for their correspondence to ARC-1 indices and concrete task descriptions and visualizations.

\noindent\textbf{Baselines:} We include 6 baselines in the comparison experiments: 1) \emph{Ours}: the \emph{Qwen-7B} model evolved by our ERO on the given ARC-1 reasoning task instance; 2) \emph{Qwen-7B}: the same \emph{Qwen-7B} pre-trained checkpoint, without ERO's evolution; 3) \emph{Qwen-32B}\footnote{\textcolor{blue}{https://dashscope.aliyuncs.com/compatible-mode/v1}}: a much larger \emph{Qwen} model with stronger general task solving ability than the 7B model; 4) \emph{GPT-4o-mini}\footnote{\textcolor{blue}{https://platform.openai.com/docs/models/gpt-4o-mini}}, 5) \emph{GPT-4o}\footnote{\textcolor{blue}{https://platform.openai.com/docs/models/gpt-4o}} and 6) \emph{GPT-5}\footnote{\textcolor{blue}{https://platform.openai.com/docs/models/gpt-5}}, which are three GPT-series models enhanced with multi-modal processing ability and chain-based reasoning capability. For \emph{Ours} and \emph{Qwen-7B}, we deploy their checkpoints at our local GPU server. For the rest of baselines, we call their corresponding APIs. Their key hyper-parameters such as temperature and top-p sampling rate follow default values. For GPT-5, we use its default reasoning efforts level~(``minimal'').

\subsection{Major Results}
For all of the selected baselines, we use a pre-designed standard prompt template to ensure fair evaluation, as illustrated in Figure~\ref{fig:prompt}. By using this standard template, we could test selected baselines on the 15 reasoning tasks sampled from ARC-1 benchmark, and then compute their per-instance pass@1 reasoning scores~(as we defined in Eq.~(\ref{eq:score})). We next present these results and corresponding discussion. 

\begin{figure}[t]
    \centering
    \includegraphics[width=0.99\columnwidth]{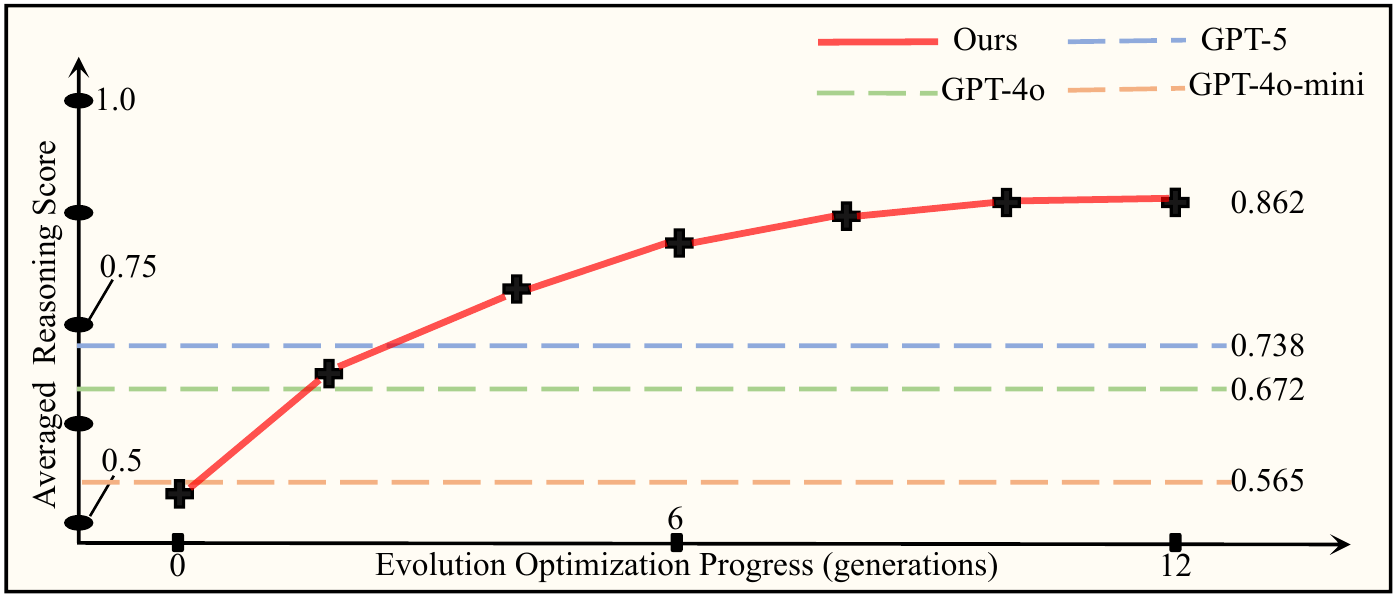}
    \caption{Evolution curve of ERO on ARC benchmark.}
    \label{fig:evolution}
\end{figure}

\noindent\textbf{Evolutionary Convergence:} We first demonstrate the effectiveness of our ERO by illustrating its evolutionary convergence curve as shown in Figure~\ref{fig:evolution}, where each scalar point in the red line is the average reasoning score of ERO across 15 tested reasoning tasks. We also attach the average scores of three advanced GPT-series baselines with dashed lines. The results  in Figure~\ref{fig:evolution} demonstrate that while a simple pre-trained Qwen-7B model underperforms the GPT models due to its limited capacity and pre-training data scale, it could be evolved by our ERO to surpass these advanced baselines on reasoning tasks. This finding may also indicates the knowledge prior redundancy of existing LLMs. We may not need continually scale both the model capacity and training data size to enable LLM's human-level reasoning ability. On the contrary, such ability may conceal itself within the LLM's parameters, and could be adapted to specific reasoning task through evolution. The results above at least demonstrate potential of evolutionary algorithms on LLM's post-tuning.

\begin{figure*}[t]
    \centering
    \includegraphics[width=0.90\linewidth]{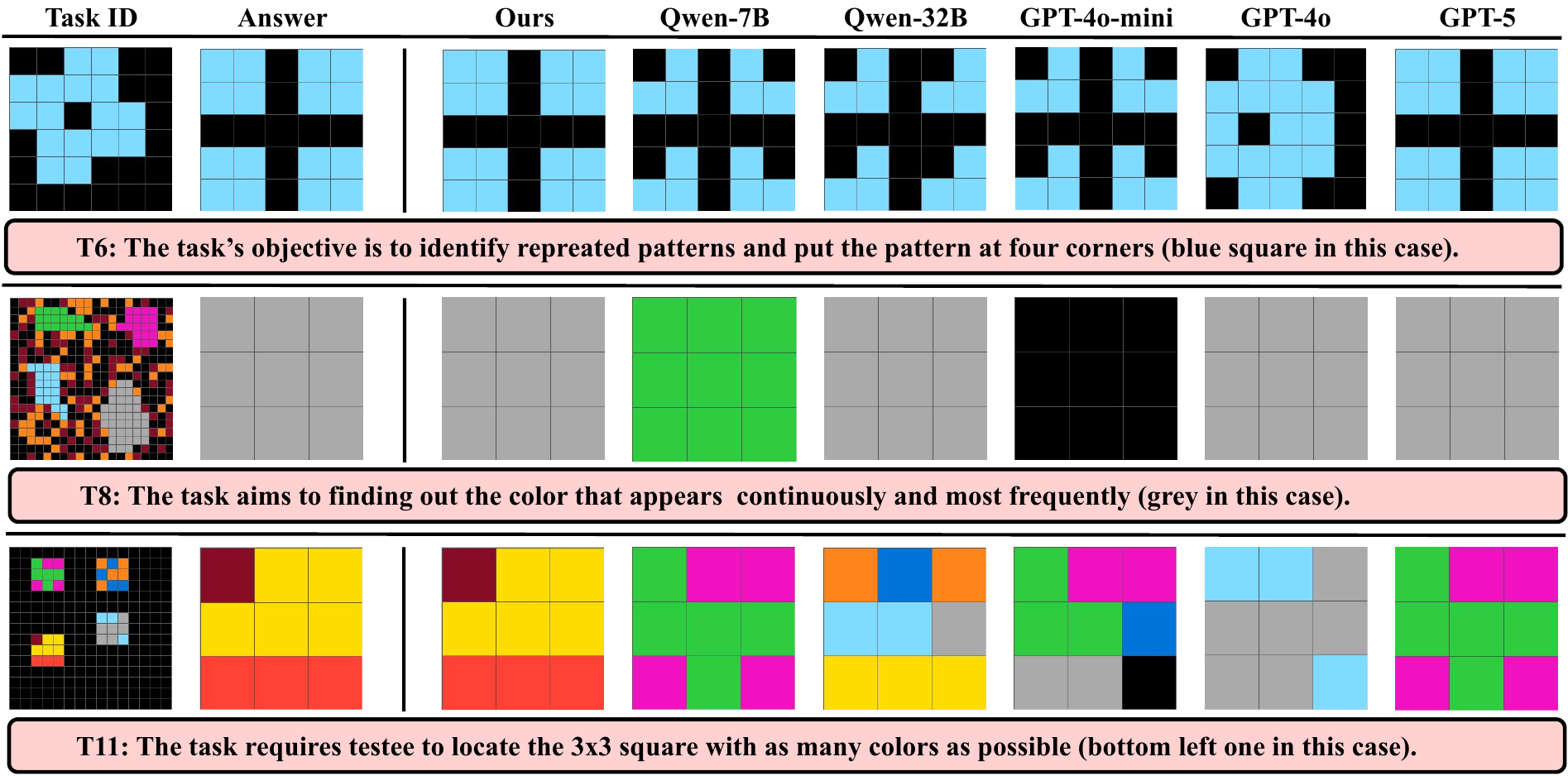}
    \caption{Showcases on the effectiveness our ERO for  boosting the understanding and reasoning ability of LLMs.}
    \label{fig:showcase}
\end{figure*}

\noindent\textbf{Performance Comparison:} We further present the per-instance performance comparison between our ERO and other baselines in the lower half of Table~\ref{tab:results}. Where the best and second-best are labeled in bold and underlined respectively. We also specifically mark the results of our ERO and Qwen-7B in light blue to highlight the relative improvement. From the results, we can observe that: 1) ERO could significantly improve the reasoning capability of Qwen-7B through 12 evolution generations, which cross-validates that intelligence (whether organic or machine-based) obeys evolution principle~(\emph{survival-of-the-fittest}); 2) With our ERO, a relatively weak Qwen-7B LLM could be evolved to perform competitively with one of the most advanced LLMs: GPT-5. On 8 of the 15 tested task instances, ERO presents significant performance advantage; 3) The reasoning capability of LLMs may not root from existing scaling law in training these LLMs. A direct evidence lies in the comparison between Qwen-7B and Qwen-32B models. On 8 of the 15 reasoning tasks, a smaller Qwen-7B model presents better logical reasoning and understanding level than its ``improved version''. This might indicate that we should pay more attention on multi-dimensional solutions for reasoning enhancement of LLMs, not only the scale of LLM pre-training.

We also showcase in Figure~\ref{fig:showcase} three task instances~(T6, T8 and T11) where our ERO successfully evolves the initial Qwen-7B model from completely wrong reasoning to crystal correct answer. As their descriptions and visualizations presented in the figure, these ARC-1 reasoning tasks challenge the innate abilities of intelligence of human beings, let alone the LLMs never being trained on such tasks. 

\subsection{An Important Future Work}
In this paper, we mainly focus on the evolution of LLM's reasoning capability under a give reasoning task. While the results mentioned in previous sections have clearly demonstrated that introducing evolutionary perspective into LLM's intelligence enhancement could result in surprising and promising effects, we have to note that the evolution of human beings may not be such simple, i.e., in an \emph{adaption-per-task} fashion. On the contrary, the subtle evolution of human beings emerges in the remix of complex environmental dynamics and concurrent multitasking. This outlines an important and promising future work of our ERO, which is the meta-evolution across reasoning task distribution:
\begin{equation}
    \mathbb{S}_{meta} = E_{\tau \sim \Omega}[\mathbb{S}(\theta|\tau)],
\end{equation}
which is the expectation of reasoning scores over a reasoning task distribution $\Omega$. As computing power and evolution paradigm~(e.g., \cite{es4llm}) continue to iterate and update, we may witness in the near future the emergence of machine intelligence species with diverse behavior and characteristics~(e.g., ``\emph{The Matrix}'' movie), purely by evolution. 

\section{Conclusion}
The position of this paper bridges the evolutionary computation community and LLMs community by proposing the ERO framework, which iteratively evolves LLM's parameters to maximize its System 2 reasoning scores on given reasoning tasks. At the core of ERO, we introduce island architecture-based evolutionary strategy to ensure searching diversity and quality, which attains reasoning performance gain effectively. Combined with specially designed cache optimization and ray acceleration mechanisms, ERO is capable of evolving a large population of LLMs on relatively limited computational resources. We validate ERO's potential by comparing it to existing representative LLMs on ARC benchmark. The promising results not only demonstrate evolution of LLMs is useful for intelligence enhancement, but may also reveal implicit connections between organic human beings and connectionism-based machine intelligence. We hope this work could appeal for more research efforts on evolutionary machine intelligence, and more importantly, exploration on more possibilities.   

\section*{Acknowledgments}
This work was supported in part by National Natural Science Foundation of China (Grant No. 62276100), in part by the Guangdong Provincial Natural Science Foundation for Outstanding Youth Team Project (Grant No. 2024B1515040010), in part by Guangzhou Science and Technology Elite Talent Leading Program for Basic and Applied Basic Research (Grant No. SL2024A04J01361), and in part by the Fundamental Research Funds for the Central Universities (Grant No. 2025ZYGXZR027). We also acknowledge that the source material we used in Figure~\ref{fig:intro} are partly designed by Freepic and Wikipedia.

\bibliography{aaai2026}

\end{document}